\documentclass[sigconf,authorversion]{acmart}
\usepackage{makecell}
\usepackage{multirow}
\usepackage{microtype}
\usepackage{colortbl}
\usepackage{diagbox}
\usepackage{graphicx}
\AtBeginDocument{%
\providecommand\BibTeX{{%
		\normalfont B\kern-0.5em{\scshape i\kern-0.25em b}\kern-0.8em\TeX}}}

\copyrightyear{2023}
\acmYear{2023}
\setcopyright{acmlicensed}\acmConference[CIKM '23]{Proceedings of the 32nd
ACM International Conference on Information and Knowledge
Management}{October 21--25, 2023}{Birmingham, United Kingdom}
\acmBooktitle{Proceedings of the 32nd ACM International Conference on
Information and Knowledge Management (CIKM '23), October 21--25, 2023,
Birmingham, United Kingdom}
\acmPrice{15.00}
\acmDOI{10.1145/3583780.3614833}
\acmISBN{979-8-4007-0124-5/23/10}

\acmSubmissionID{989}

\begin{document}

\title{DebCSE: Rethinking Unsupervised Contrastive Sentence Embedding Learning in the Debiasing Perspective}


\author{Pu Miao}
\orcid{0000-0002-9308-5247}
\authornote{All authors contributed equally to this research.}
\affiliation{%
	\institution{Sina Weibo}
	\streetaddress{West 8 Xibeiwangdong Rd}
	\city{Beijing}
	\country{China}}
\email{miaopu@staff.weibo.com}

\author{Zeyao Du}
\orcid{0000-0002-9054-3337}
\affiliation{%
	\institution{China Literature Limited}
	\streetaddress{Binjiangzhongxin MT}
	\city{Shang Hai}
	\country{China}}
\email{duzeyao@yuewen.com}

\author{Junlin Zhang}
\orcid{0000-0002-7982-3194}
\affiliation{%
	\institution{Sina Weibo}
	\streetaddress{West 8 Xibeiwangdong Rd}
	\city{Beijing}
	\country{China}}
\email{junlin6@staff.weibo.com}

\renewcommand{\shortauthors}{PuMiao,Zeyao Du and Junlin Zhang}
\begin{abstract}
	Several prior studies have suggested that word frequency biases can cause the Bert model to learn indistinguishable sentence embeddings. Contrastive learning schemes such as SimCSE and ConSERT have already been adopted successfully in unsupervised sentence embedding to improve the quality of embeddings by reducing this bias. However, these methods still introduce new biases such as sentence length bias and false negative sample bias, that hinders model's ability to learn more fine-grained semantics.
	In this paper, we reexamine the challenges of contrastive sentence embedding learning from a debiasing perspective and argue that effectively eliminating the influence of various biases is crucial for learning high-quality sentence embeddings. We think all those biases are introduced by simple rules for constructing training data in contrastive learning and the key for contrastive learning sentence embedding is to "mimic" the distribution of training data in supervised machine learning in unsupervised way. We propose a novel contrastive framework for sentence embedding, termed DebCSE, which can eliminate the impact of these biases by an inverse propensity weighted sampling method to select high-quality positive and negative pairs according to both the surface and semantic similarity between sentences. Extensive experiments on semantic textual similarity (STS) benchmarks reveal that DebCSE significantly outperforms the latest state-of-the-art models with an average Spearman’s correlation coefficient of 80.33\% on BERTbase. 
\end{abstract}

\ccsdesc[500]{Computing methodologies~Information extraction}
\keywords{Sentences Embeddings, Contrastive Learning, Information Extraction, Text Retrieval, Natural Language Understanding}



\maketitle

\section{Introduction}

Sentence representation learning is one of the most important tasks of natural language processing (NLP), and high-quality sentence embedding is useful for a large number of downstream tasks, such as semantic textual similarity measure and semantic retrieval. Though pre-trained language model \cite{kenton2019bert,yang2019xlnet} achieves great successes in many  supervised tasks, it actually damages the quality of sentence embeddings \cite{jiang2022promptbert}. Sometimes it even underperforms the traditional approaches such as GloVe \cite{pennington2014glove}. 

Some previous work\cite{li2020sentence} has attributed this kind of poor performance to the word frequency bias introduced by masked language modeling (MLM) pre-training task in BERT\cite{devlin2018bert} or RoBERTa \cite{liu2019roberta}. For example, Bert-Flow \cite{li2020sentence} find out that the masked language modeling task makes the embedding of high-frequency words concentrate densely while low-frequency words disperse sparsely, which leads to an anisotropy word embedding space and results in a high similarity between any sentence pair. 

Recent works \cite{gao2021simcse,yan-etal-2021-consert} have shown that contrastive learning helps pre-trained language models to reduce the word frequency bias and learn good sentence embeddings without any labeled data. Contrastive learning aims to generate a representation that maximizes the similarity between positive pairs while minimizing the similarity between negative pairs. As a representative method, unsupervised SimCSE\cite{gao2021simcse} passes the same input sentence to the pre-trained encoder twice and obtain two embeddings as “positive pairs”. It randomly selects some other sentences in the same batch as “negative pairs”. Even though SimCSE is rather simple, sentence embeddings learned by it have been shown to be better than other more complicated methods\cite{gao2021simcse}.

\begin{figure*}[htbp]
	\begin{minipage}[t]{0.5\linewidth}
		\centering
		\includegraphics[width=\textwidth]{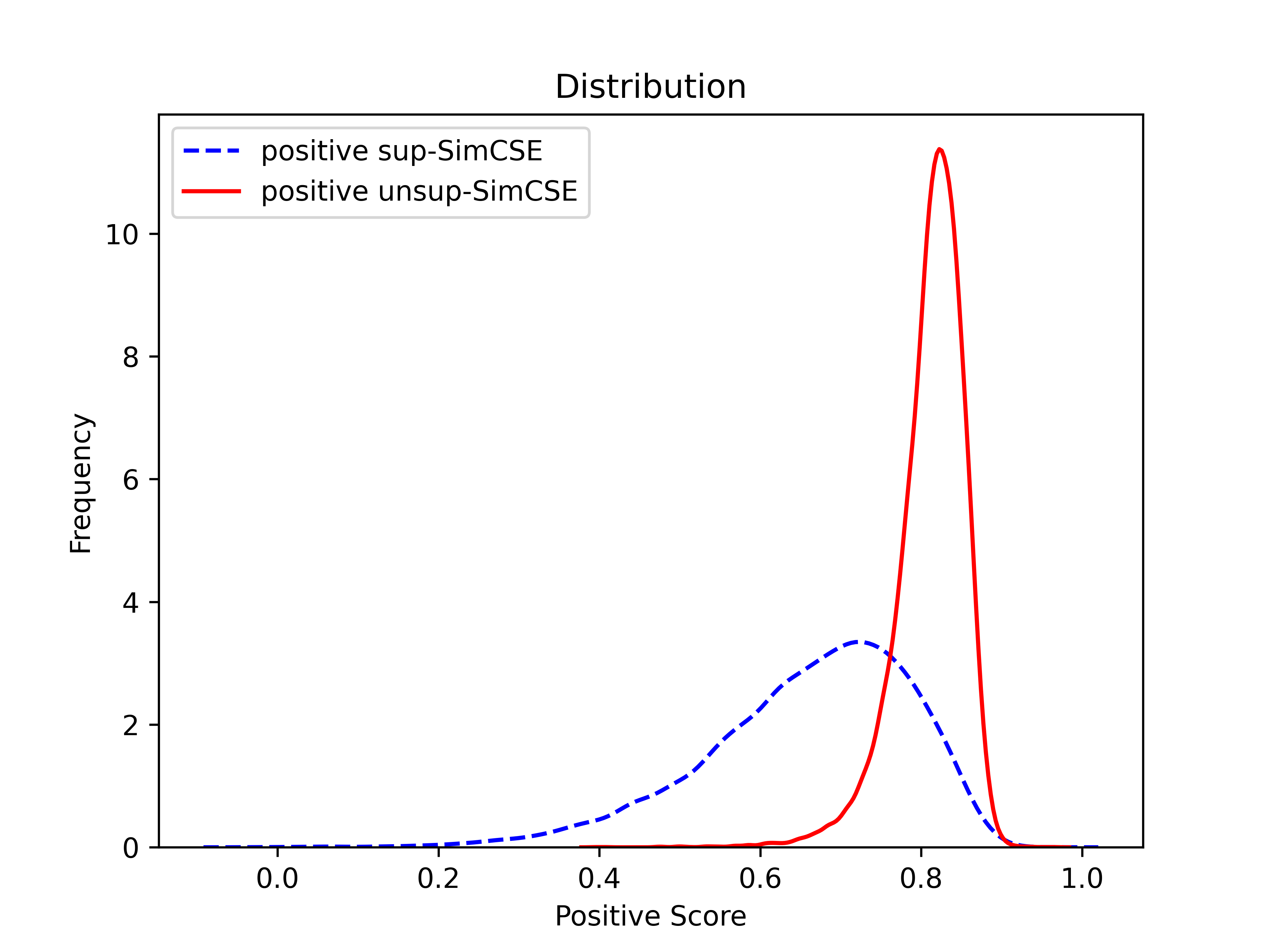}
		\centerline{(a) Semantic Similarity distribution of positive pairs }
	\end{minipage}%
	\begin{minipage}[t]{0.5\linewidth}
		\centering
		\includegraphics[width=\textwidth]{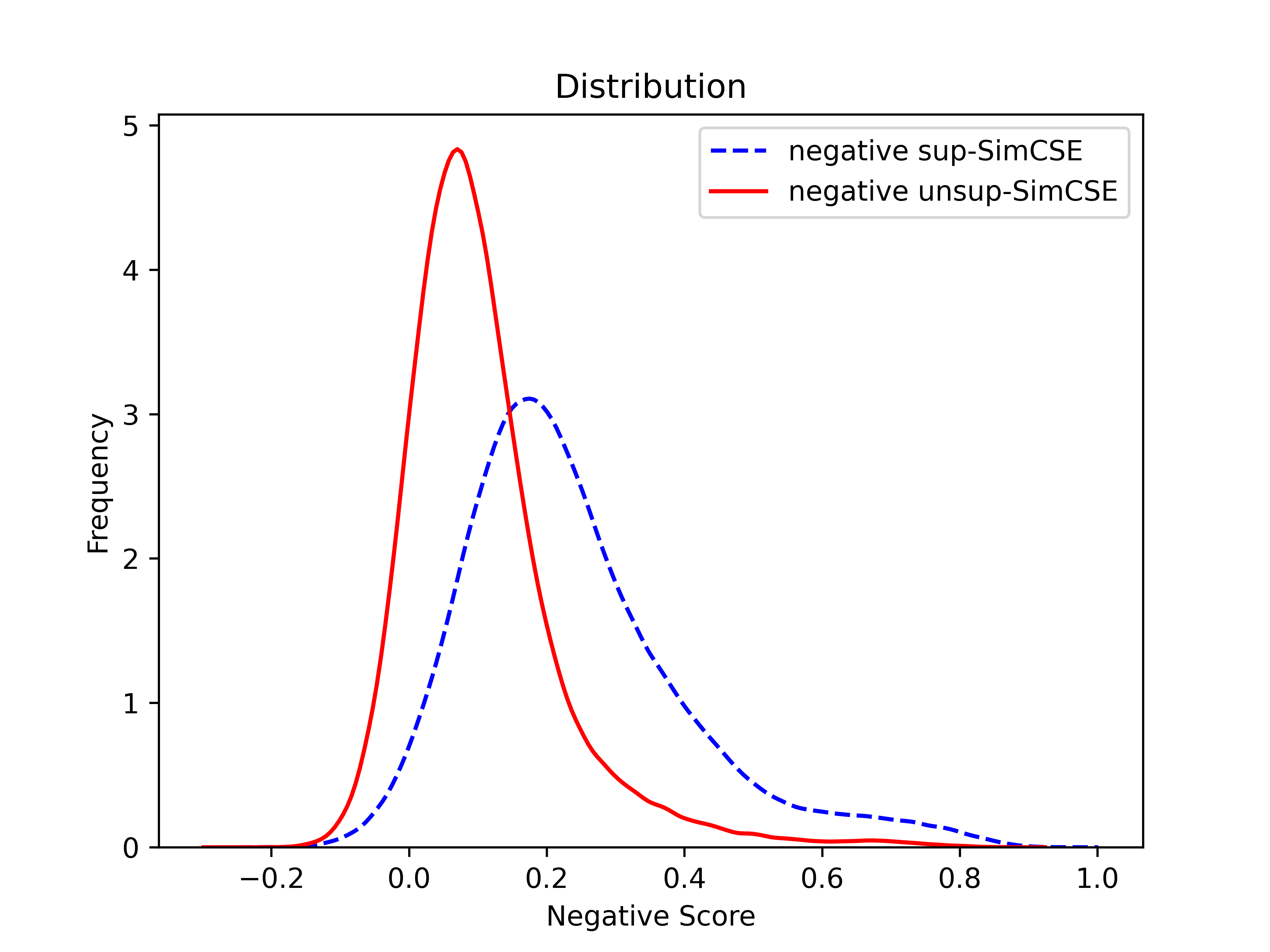}
		\centerline{(b) Semantic Similarity distribution of negative pairs}
	\end{minipage}
	\caption{Training Data Distribution Bias(Semantic) }
	\label{figure1}
\end{figure*}

Most of the following works \cite{wu2021esimcse,chuang2022diffcse,wang2022improving} adopt the similar strategy as SimCSE to construct the positive and negative pairs. However, we argue that this kind of strategy is too simple and brings new types of bias, which degrades the quality of sentence embedding. Some other studies also find the problem and try to solve this problem by reducing some specific kind of bias. For example, ESimCSE\cite{wu2021esimcse} argues SimCSE contains a "sentence length bias" because positive pair has the same length information, which has negative effect on sentence representation. The authors applies a simple word repetition operation to modify sentence lengths in a positive pair to alleviate this problem. Another known bias is called "false negative sample bias"\cite{chuang2020debiased}, which means sampling negative pairs at random will inevitably draws some “false negatives”. The authors propose an instance weighting method to resolve this problem.

Compared with the complex distribution of positive and negative pairs in supervised machine learning models, the training data constructed by simple rules in contrastive learning contains various bias, known or unknown. We think all those biases are introduced by these simple rules. Obviously, there are huge differences in similarity distribution of training data between contrastive learning and supervised machine learning, which can be regarded as a kind of "training data distribution bias" for constrastive learning. In order to reduce this bias, we argue that the key for contrastive learning sentence embedding is to "mimic" the distribution of training data in supervised machine learning in unsupervised way, which can be regarded as a kind of debiasing procedure.

To tackle this problem, in this paper we propose a simple yet effective framework DebCSE, which regards the contrastive learning for unsupervised sentence representation as a debiasing process to take both the known biases, such as sentence length bias and false negative sample bias, and some unknown biases into account. Specifically, given an input sentence, we generate the positive candidates and sample the positive pairs from candidates by using an inverse propensity weighted sampling method, which opts to select the candidate  with low surface similarity and high semantic similarity with input sentence as the positive example. For negative pairs, we filter out candidates according to embedding similarity and then also leverage the inverse propensity weighted sampling method to select high-quality negatives which have high surface similarity and low semantic similarity with input sentence.
Besides, we also propose an alternative normalization contrastive loss in this paper to enforce the sentence embedding to be close to its sampled positive pairs, and
far apart from the sampled negative pairs as well as the random in-batch negative pairs in an alternative way.

Extensive experimental results on semantic textual similarity tasks demonstrate the effectiveness of our proposed model compared to strong baselines. For example, DebCSE achieves on average 4.1 points absolute improvements in terms of Spearman’s correlation on BERTbase compared with SimCSE. Our experimental results also show that the DebCSE model effectively alleviates the above-mentioned "training data distribution bias".

Our contributions are summarized as follows:

(1) To our knowledge, our approach is the first attempt to reexamine the challenges of contrastive sentence embedding learning from a debiasing perspective. We argue that reducing the "training data distribution bias" is the key problem for better sentence embedding.

(2) We propose DebCSE, a debiased contrastive learning framework that takes both the surface similarity and semantic similarity between sentences into account and incorporates inverse propensity weighting  method to sample high-quality negatives and positives, which guarantee the representation quality of the sentence embedding.

(3) Experimental results on seven semantic textual similarity tasks show that DebCSE outperforms the baselines by a substantial margin.

\section{Biases in Contrastive Sentence Embedding}

In the following, we summarize the biases in contrastive sentence embedding. First, we describe the known biases mentioned in the literature such as word frequency bias, sentence length bias and false negative sample bias. Then we prove the existence of the training data distribution gap between the contrastive learning and supervised learning, which is called "training data distribution bias" in this paper.

\subsection{Word Frequency Bias in Sentence Embedding}

Studies in \cite{gao2018representation} have proved that most of the learnt word embedding tend to degenerate and be distributed into a narrow cone when training a model for natural language generation tasks through likelihood maximization. This representation degeneration problem limits the representation power of word embedding and the learned model parameters do not have enough capacity to represent the diverse semantics in natural language. They also pointed out that this anisotropy is highly relevant to the imbalance of word frequency.

Studies in \cite{li2020sentence} verify this hypothesis in the context of BERT and indicate that the word embedding can be biased to word frequency. They observe that high-frequency words concentrates densely and low-frequency words disperse sparsely in the learned anisotropic embedding space. Therefore, both the resulted sentence embedding and the induced sentence similarity are problematic.

\subsection{Biases in Contrastive Sentence Embedding Learning}

Some following  methods such as SimCSE\cite{gao2021simcse} and ConSERT \cite{yan-etal-2021-consert} introduce the contastive learning into the unsupervised sentence embedding task and alleviate the word frequency bias to some extent.

As a current state-of-the-art method, unsupervised SimCSE feeds each sentence to the pre-trained BERT twice with two independently sampled dropout masks to produce positive pair while randomly chose two sentences in a batch as negative pair. It's obvious that a positive pair would contain the same length information because it's derived from the same sentence while two different sentences in negative pair usually have different length information. Therefore, the different length information contained by positive pairs and negative pairs can act as a strong feature to distinguish them. The model trained with these pairs contains "sentence length bias", which opt to consider that two sentences with similar lengths have near distance in semantics space. ESimCSE\cite{wu2021esimcse} applies a simple word repetition operation to modify  sentence lengths in a positive pare to alleviate the "sentence length bias".

While significant improvement is achieved,  contrastive learning approaches generally do not take into account the high-level semantics contained in images or sentences. For example, most contrastive learning frameworks, both in CV and NLP, directly regards other samples within the same training batch as negatives without considering the semantics of the content. Obviously, this kind of sampling negative pairs at random will inevitably draws some “false negatives” and learn less effective representation models. To alleviate the above "false negative sample bias" problem, some work\cite{chuang2020debiased} proposes an instance weighting method to punish false negatives and generate noise-based negatives to guarantee the uniformity of the representation space.

\subsection{Training Data Distribution Bias}
Most current contrastive sentence learning methods leverage either same input sentences with different dropout ratio as positive pairs or random in-batch samples as negative pairs. So the training data construction procedure of contrastive sentence learning can be regarded as a data generator consisting of some simple heuristic rules. However, these simple training data construction rules bring known or unknown new bias as discussed above. In contrast to contrastive sentence learning, the training data of supervised learning is usually produced by human labor and shows high quality to train better models without bias introduced by simple rules. Many models using supervised datasets generally outperform self-supervised approach\cite{gao2021simcse,chuang2020debiased,yan-etal-2021-consert} while they need extensive human labour for the ground truth labels. 

In this section, we will empirically prove that there is huge training data distribution gap between the contrastive sentence learning and supervised learning methods. By taking both the surface and semantic similarity into consideration, we extract the training data of two different approaches to show the distribution gap between them.

The supervised natural language inference (NLI) datasets  have been proved to be effective for learning sentence embeddings, by predicting whether the relationship between two sentences is entailment, neutral or contradiction. In NLI datasets, given one premise, annotators manually write one entailment sentence that is absolutely true , one neutral sentence that might be true, and one contradiction sentence that is definitely false. Therefore, we use entailment pairs and its contradiction pairs from the NLI datasets as supervised positive pairs and negative pairs, respectively.  

For the training data of contrastive sentence learning, we follow SimCSE and use the same sentence augmented by different dropout ratio as positive pairs and random in-batch examples as negative pairs.

\begin{table}[htbp]
	\centering
	\caption{Training Data Distribution Bias (Surface)}
	\label{table1}%
	\begin{tabular}{ccc}
		\toprule
		\textbf{} & {\textbf{Supervised}} & {\textbf{Contrastive}} \\
		\midrule
		
		Positive & 0.38  & 1.0\\
		Negative  & 0.28   & 0.09\\
		\bottomrule
	\end{tabular}
\end{table}%

After obtaining the training data, we firstly use the lexical overlap ( the ratio of shared word number to the longer sentence length between two sentences ) between the positive pairs or negative pairs as surface similarity of training data. Table {1} shows the distribution difference between contrastive and supervised approaches. It can be seen that contrastive method has a much higher surface similarity for positive pairs and  much lower surface similarity for negative pairs, compared with the training data distribution of supervised learning. This demonstrates that positive pairs are too similar and negative pairs have little common words for contrastive learning, which means the training data is too easy for model to learn and contains little useful information.

Then, we use a pre-trained SimCSE as sentence encoder and cosine function as similarity metric to measure semantic similarity of positive and negative pairs for both contrastive and supervised learning. Figure 1 shows the results. We can see that contrastive method has higher semantic similarity for positive pairs and lower semantic similarity for negative pairs, compared with the training data distribution of supervised learning. This also indicates the training data of contrastive sentence learning is too easy and uninformative even in semantic perspective.
\begin{figure}[h]
  \centering
 
  \includegraphics[width=1.0\linewidth]{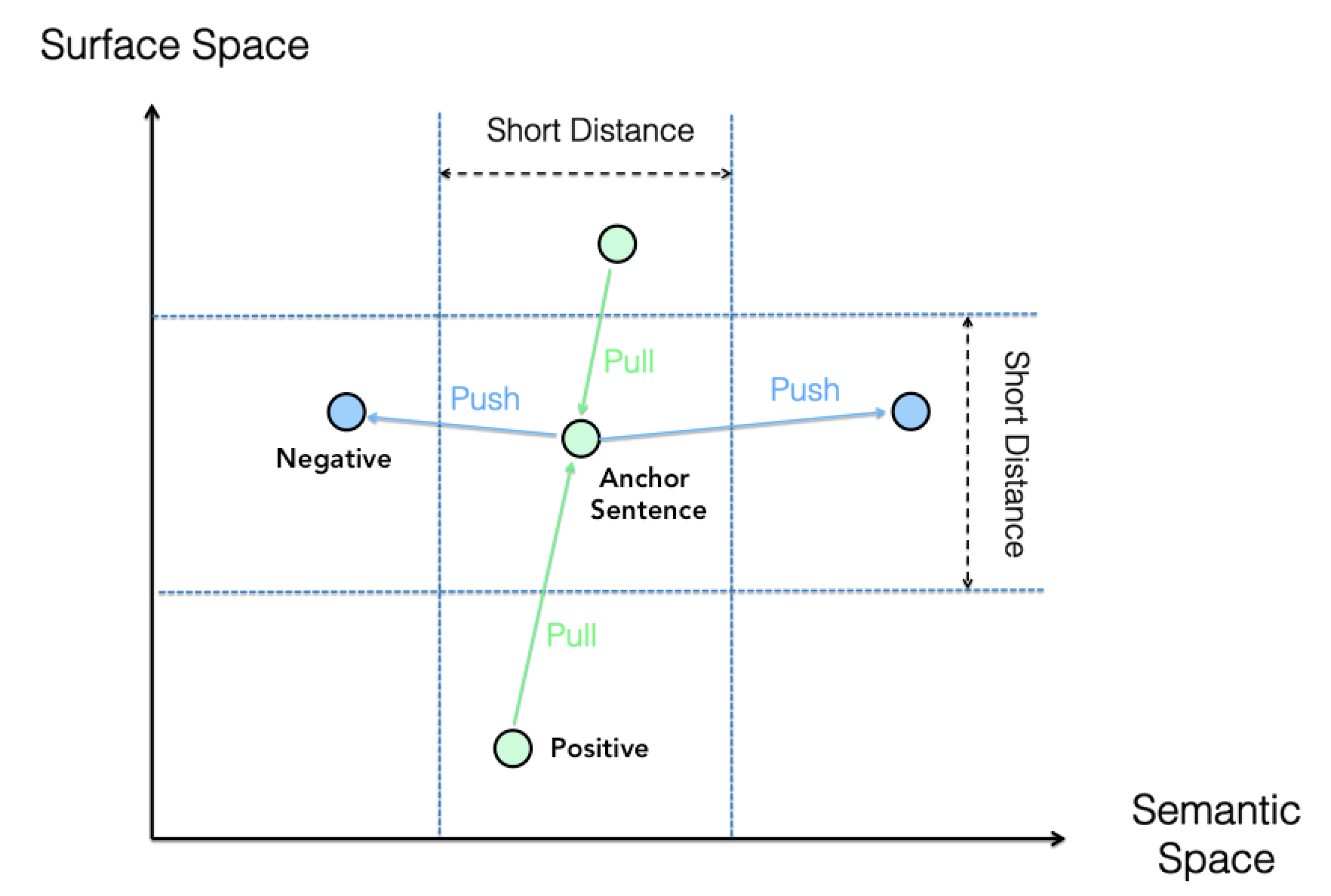}
  \caption{In both semantic space and surface space, good positive examples should have close semantic distance but distant surface distance from the anchor sentence, while good negative examples should have distant semantic distance but close surface distance from the anchor sentence.}
  \Description{}
  \label{Fig.semantic}
\end{figure}

In this paper, we argue that it's the simple training data construction rules adopted in contrastive sentence learning that introduce many biases into sentence embedding and have huge distribution gap with training data used in supervised learning. How to fill this distribution gap is key to a high quality sentence embedding. We call this distribution gap between the contrastive and supervised learning "training data distribution bias" in this paper.

\begin{figure*}[h]
	\centering
	\includegraphics[width=0.95\linewidth]{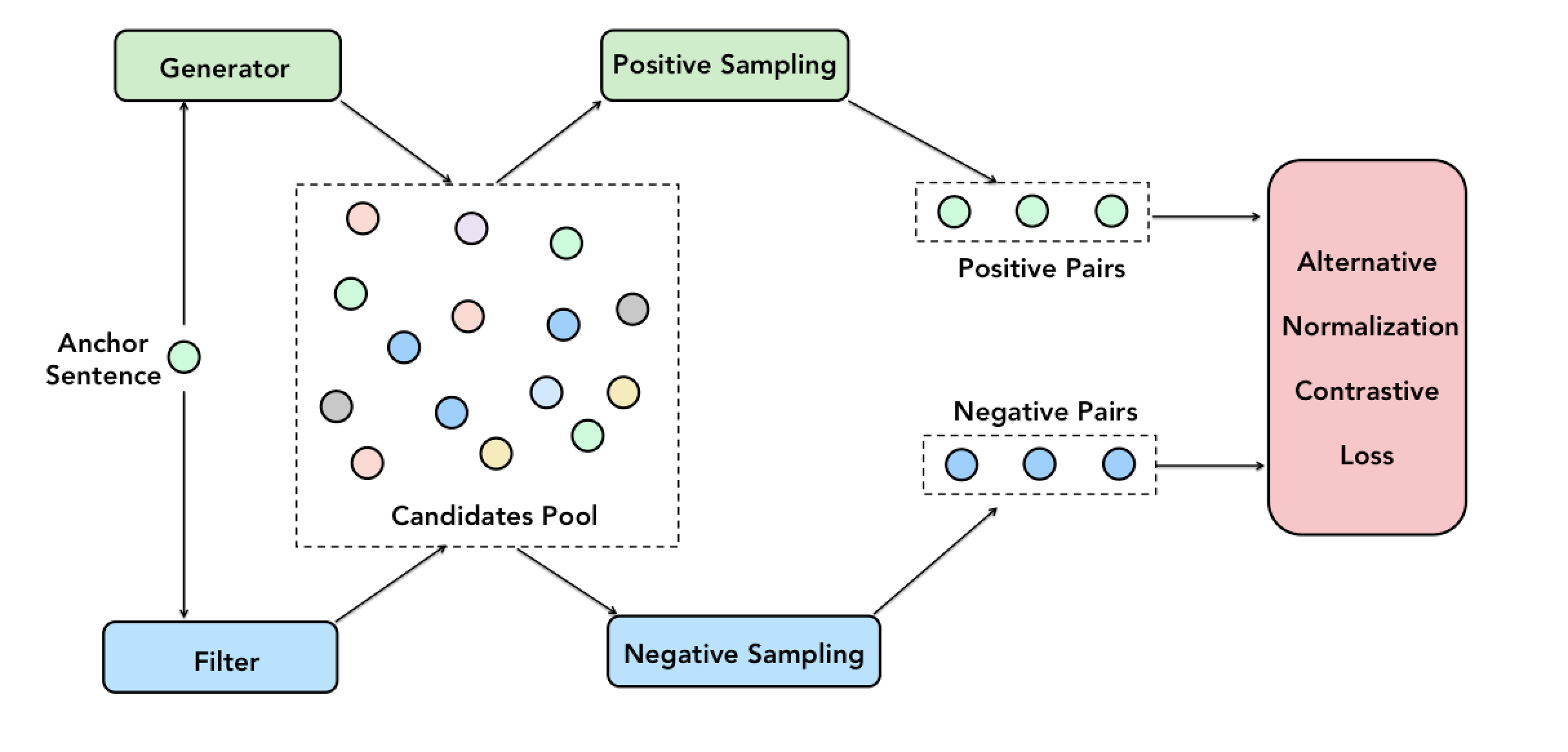}
	\caption{The overall architecture of DebCSE}
	\label{figure3}
\end{figure*}

\section{Problem Formulation}
In this section, we specify the research problem of unsupervised contrastive sentence embedding learning in the debiasing perspective. As discussed above, the training data distribution gap between contrastive and supervised learning hinders the success of unsupervised contrastive sentence embedding learning. 

To tackle this problem, we propose to fill the distribution gap as a debiasing task by "mimicking" the training data distribution of supervised learning in an unsupervised way, which allows our proposed approach to possess both the low-cost advantage of contrastive learning and avoid various biases in training data it may introduce. Specifically, we take both the surface and semantic similarity of positive and negative pairs into account when we automatically construct training data of contrastive learning. Our basic assumption is as follows: Given the input sentence $x_i$, an informative positive example $x_i^+$ should have close semantic distance in semantic space and distant surface distance in surface space with $x_i$ simultaneously while a good negative example $x_i^-$  should have  close surface distance but distant semantic distance with $x_i$ at the same time. Figure 2 shows our proposed assumption.

Following this basic assumption of constructing training data, we can produce candidates pool $C_{x_i}=\{x_j,j<n\}$ for sentence $x_i$ in unsupervised way and sample positive or negative pairs from  $C_{x_i}$ by inverse propensity weighting probability which can be computed according to the basic assumption. 

Inverse propensity weighting is a widely used debiasing technology in recommendation systems \cite{ai2018unbiased} \cite{wang2016learning} \cite{joachims2017unbiased}, which is unbiased for any probabilistic assignment mechanism and leverages the inverse of the propensity score (observed item clicked probability in recommendation system) to align the biased item distribution over the click space with that over the exposure space. Similarly, we use inverse propensity weighting to align the biased training data distribution over the contrastive learning with that over the supervised learning. By applying this alignment, we can effectively mitigate the disparities in data distribution between the two approaches. However, in this paper we leverage inverse propensity weighting score as sampling probability to sample positive or negative pairs from candidates pool $C_{x_i}=\{x_j,j<n\}$, which is different from the common usage in recommendation systems.

\section{Our Proposed Method}
In this section, we present our DebCSE to effectively alleviate the training data distribution bias by negative and positive sampling. The overall architecture of DebCSE is illustrated in Figure 3, consisting of three main components: negative sampling by inverse propensity weighting (Section 4.1), positive sampling by inverse propensity weighting (Section 4.2), and alternative normalization contrastive loss (Section 4.3).

\subsection{Negative Sampling by Inverse Propensity Weighting Probability}
In this section, we describe how to sample negative pairs by inverse propensity weighting probability, which consists of two phases: First, we filter out negative candidates to form the negative candidates pool based on the input sentence; Then, we sample high-quality negatives according to the sampling probability.

\paragraph{\textbf{Negative Candidates Pool}}
Given an input sentence $x_i$, we firstly form the negative candidates pool $C_{x_i}^{NCP}=\{x_j,j<n\}$  by filtering out sentences $x_j$ with semantic similarity $Sim(x_i,x_j)$ within specific range. Specifically, letting $h_i$ denotes sentence embedding of input sentence $x_i$, we firstly generate sentence embeddings for sentences in dataset using the trained encoder such as unsupervised SimCSE model.
Then, for any sentence $x_i$ in dataset, we leverage $cos(h_i,h_j)$ as semantic similarity metric to find out those $x_j$ which have semantic similarity with $x_i$ within specific range. We filter out those sentences with not too high or not too low semantic similarity with $x_i$ as candidate negative pairs. For those $x_j$ with too low semantic similarity with $x_i$, they are too easy negative examples just like random negatives containing little knowledge for model to learn. For those $x_j$ with too high semantic similarity with $x_i$, they have much higher probability of being the “false negatives” introducing noise into the model. Therefore, we ignore these examples and pay more attention on much informative examples.

\paragraph{\textbf{Negative Sampling}}
Given the input sentence $x_i$ and its corresponding negative candidates pool  $C_{x_i}^{NCP}$, we opt to select few high quality negatives among them because the candidate number is still too large . One straightforward method is to rank them according to semantic similarity with $x_i$ and select top ranked sentences as negative examples for $x_i$. However, higher semantic similarity means higher "false negatives" probability.

To resolve this problem, we leverage inverse propensity weighting probability as sampling probability by taking both sentence surface similarity and semantic similarity into account. Specifically, edit distance \cite{ristad1998learning} $edit(x_i,x_j)$ and embedding cosine function $cos(h_i,h_j)$ are used as similarity metrics to measure the surface dissimilarity and semantic similarity of $x_j$  with $x_i$, respectively. To make the two similarity score comparable, we use $Softmax(\cdot)$ function to normalize each score as follows:
\begin{equation}
	S_{sur}(x_i,x_j) =  1-\frac{e^{edit(x_i,x_j)} } {\sum_{k = 1}^{n}  e^{edit(x_i,x_k)}} 
\end{equation}
and
\begin{equation}
	S_{sem}(x_i,x_j) =  \frac{e^{cos(x_i,x_j)} } {\sum_{k = 1}^{n}  e^{cos(x_i,x_k)}} 
\end{equation}
where $n$ is the number of negative examples in the negative candidates pool $C_{x_i}^{NCP}$, while $S_{sur}(x_i,x_j)$ and $S_{sem}(x_i,x_j)$ are surface similarity and semantic similarity scores, respectively.

In order to find out those $x_j$ with higher surface similarity and lower semantic similarity with $x_i$ simultaneously, we take both the surface and semantic similarity into consideration and propose to sample the negative examples according to the following inverse propensity weighting probability:
\begin{equation}
	\Tilde{P}_{neg}(x_j|x_i) =  (1-\lambda_n)S_{sur}(x_i,x_j)+\lambda_n(1-S_{sem}(x_i,x_j))
\end{equation}
where $ \lambda_n$ is the balanced hyper-parameter for adjusting the importance of semantic similarity. 

To make the model pay more attention to more informative and better quality negative examples among the candidates during the sampling procedure, $Softmax(\cdot)$ function is used to sharpen the sampling probability as follows:
\begin{equation}
	P_{neg}(x_j|x_i) =   \frac{e^{\Tilde{P}_{neg}(x_j|x_i)} } {\sum_{k = 1}^{n}  e^{\Tilde{P}_{neg}(x_k|x_i)}} 
\end{equation}

Then, we sample from the pool of $n$ negative candidates, each with a probability of ${P}(x_j|x_i)$, according to the rules that higher probability to be selected as negative of $x_i$ for those examples with higher sampling probability. As a result, we obtain a set $C_{x_i}^{neg}=\{ x_1,x_2....,x_m\}$ consisting of $m$ sampled negative examples for input sentence $x_i$. Here we let $m=2$ in our experiments.

\subsection{Positive Sampling by Inverse Propensity Weighting Probability}

In this section, we describe how to sample positive pairs by inverse propensity weighting probability, which also consists of two phases: First, we generate positive candidates to form the positive candidates pool based on the input sentence; Then, high-quality positives are sampled according to the sampling probability.

\paragraph{\textbf{Positive Candidates Pool}}
Given the input sentence $x_i$, we try to generate augmented positive candidate samples exhibiting certain kinds of variation in an unsupervised way. We can leverage strong text generators such as ChatGPT to automatically produce positive candidates through in context learning prompt. However, we intent to show that a weaker text generator also works.  To this end, we utilize  back-translation model$\footnote{Opus-mt \url{https://huggingface.co/Helsinki-NLP/opus-mt-en-zh}}$  $\footnote{Opus-mt \url{https://huggingface.co/Helsinki-NLP/opus-mt-zh-en}}$ or summary model as the text generator. To produce as many different candidates as possible, we inject one or two high frequency words as noise or mask one or two words on random position in the sentence $x_i$ before sending the input sentence into text generator. It should be noticed that if ChatGPT is used as a tool for generating positive examples, our proposed model should achieve much better performance. We will leave this kind of adopting large language model to produce training data of contrastive learning as our future work.

\paragraph{\textbf{Positive Sampling}}
After obtaining corresponding positive candidates pool  $C_{x_i}^{PCP}=\{x_j,j<n\}$ of the input sentence $x_i$ , we also use inverse propensity weighting probability as sampling probability to select informative positive examples in an unsupervised manner, just like we do in negative sampling stage. However, we intent to regard those $x_j$ with lower surface similarity and higher semantic similarity with $x_i$ simultaneously as high quality positive examples. Therefore, we propose to sample the positive examples according to the following inverse propensity weighting probability:
\begin{equation}
	\Tilde{P}_{pos}(x_j|x_i) =  (1-\lambda_p)(1-S_{sur}(x_i,x_j))+\lambda_pS_{sem}(x_i,x_j)
\end{equation}
where $ \lambda_p$ is the balanced hyper-parameter. 

We also leverage $Softmax(\cdot)$ to let model focus on more informative positive examples as follows:
\begin{equation}
	{P_{pos}}(x_j|x_i) =   \frac{e^{\Tilde{P}_{pos}(x_j|x_i)} } {\sum_{k = 1}^{n}  e^{\Tilde{P}_{pos}(x_k|x_i)}} 
\end{equation}
We can sample positive examples according to the above probability to obtain the set $C_{x_i}^{pos}=\{ x_1,x_2....,x_m\}$ consisting of $m$ sampled positive examples for input sentence $x_i$. And we also sample $m=2$ positive examples for each input sentence in our experiments.

\begin{table*}[htbp]
	\centering
	\caption{Sentence representation performance on the STS tasks. We employ our method to BERT and RoBERTa in both base and large versions and report Spearman’s correlation.}
	\label{table2}%
	\resizebox{\linewidth}{!}{
		\begin{tabular}{cccccccccc}
			\toprule
			& \textbf{Models}  &  \textbf{STS12}  &  \textbf{STS13}  &  \textbf{STS14}  &  \textbf{STS15}  &  \textbf{STS16}  &  \textbf{STS-B}  &  \textbf{SICK-R}  &  \textbf{Avg.}  \\
			\midrule
			\multirow{11}{*}{BERT$\rm{_{base}}$} & BERT (first-last avg.)  & 39.7  & 59.38 & 49.67 & 66.03 & 66.19 & 53.87 & 62.06 & 56.7 \\
			& BERT-flow  & 58.4  & 67.1  & 60.85 & 75.16 & 71.22 & 68.66 & 64.47 & 66.55 \\
			& BERT-whitening  & 57.83 & 57.83 & 60.9  & 75.08 & 71.31 & 68.24 & 63.73 & 66.28 \\
			& IS-BERT  & 56.77 & 69.24 & 61.21 & 75.23 & 70.16 & 69.21 & 64.25 & 66.58 \\
			& ConSERT  & 64.64 & 78.49 & 69.07 & 79.72 & 75.95 & 73.97 & 67.31 & 72.74 \\
			& SimCSE  & 68.4  & 82.41 & 74.38 & 80.91 & 78.56 & 76.85 & 72.23 & 76.25 \\
			& ESimCSE  & 73.4  & 83.27 & 77.25 & 82.66 & 78.81 & 80.17 & 72.3  & 78.27 \\
			& SNCSE  & 70.67 & \textbf{84.79} & 76.99 & 83.69 & 80.51 & 81.35 & \textbf{74.77} & 78.97 \\
			& DiffCSE  & 72.28 & 84.43 & 76.47 & 83.9  & 80.54 & 80.59 & 71.23 & 78.49 \\
			& ArcCSE  & 72.08 & 84.27 & 76.25 & 82.32 & 79.54 & 79.92 & 72.39 & 78.11 \\
			& DebCSE  & \textbf{76.15} & 84.67 & \textbf{78.91} & \textbf{85.41} & \textbf{80.55} & \textbf{82.99} & 73.6  & \textbf{80.33} \\
			\midrule
			\multirow{7}{*}{RoBERTa$\rm{_{base}}$} & RoBERTa (first-last avg.)  & 40.88 & 58.74 & 49.07 & 65.63 & 61.48 & 58.55 & 61.63 & 56.57 \\
			& RoBERTa-whitening  & 46.99 & 63.24 & 57.23 & 71.36 & 68.99 & 61.36 & 62.91 & 61.73 \\
			& SimCSE & 70.16 & 75.19 & 65.52 & 75.08 & 71.31 & 68.24 & 63.73 & 76.57 \\
			& DiffCSE  & 70.05 & 83.43 & 75.49 & 82.81 & \textbf{82.12} & 82.38 & 71.19 & 78.21 \\
			& ESimCSE  & 69.9  & 82.5  & 74.68 & 83.19 & 80.3  & 80.99 & 70.54 & 77.44 \\
			& SNCSE  & 70.62 & 84.42 & 77.24 & 84.85 & 81.49 & 83.07 & 72.92 & 79.23 \\
			& DebCSE  & \textbf{74.29} & \textbf{85.54} & \textbf{79.46} & \textbf{85.68} & 81.2  & \textbf{83.96} & \textbf{74.04} & \textbf{80.60} \\
			\midrule
			\multirow{6}{*}{BERT$\rm{_{large}}$} & ConSERT  & 70.69 & 82.96 & 74.13 & 82.78 & 76.66 & 77.53 & 70.37 & 76.45 \\
			& unsup-SimCSE  & 70.88 & 84.16 & 76.43 & 84.5  & 79.76 & 79.26 & 73.88 & 78.41 \\
			& ArcCSE  & 73.17 & 86.19 & 77.9  & 84.97 & 79.43 & 80.45 & 73.5  & 79.37 \\
			& ESimCSE  & 73.21 & 85.37 & 77.73 & 84.3  & 78.92 & 80.73 & 74.89 & 79.31 \\
			& SNCSE & 71.94 & 86.66 & 78.84 & 85.74 & 80.72 & 82.29 & \textbf{75.11} & 80.19 \\
			& DebCSE   & \textbf{76.82} & \textbf{86.36} & \textbf{79.81} & \textbf{85.8}  & \textbf{80.83} & \textbf{83.45} & 74.67 & \textbf{81.11} \\
			\midrule
			\multirow{4}{*}{RoBERTa$\rm{_{large}}$} & unsup-SimCSE & 70.88 & 84.16 & 76.43 & 84.5  & 79.76 & 79.26 & 73.88 & 78.90 \\
			& ESimCSE  & 73.20 & 84.93 & 76.88 & 84.86 & 81.21 & 82.79 & 72.27 & 79.45 \\
			& SNCSE   & 73.71 & 86.73 & 80.35 & \textbf{86.8}  & 83.06 & 84.31 & \textbf{77.43} & 81.77 \\
			& DebCSE  & \textbf{77.68} & \textbf{87.17} & \textbf{80.53} & 85.90  & \textbf{83.57} & \textbf{85.36} & 73.89 & \textbf{82.01} \\
			\bottomrule
		\end{tabular}
	}

\end{table*}%

\subsection{Alternative Normalization Contrastive Loss}
Recently, a few methods in image contrastive learning such as BYOL \cite{grill2020bootstrap} and SimSiam \cite{chen2021exploring} have proved that batch-wise or feature-wise normalization helps  produce high-quality representations. However, we find direct normalization on sentence embedding is harmful to good text representation. Therefore, in this paper we propose an alternative normalization contrastive loss, which is a kind of indirect normalization approach. More precisely, we compute an normalized sentence embedding from one sentence in the positive pair and predict this normalized representation from another sentence in positive pair.

Given an input sentence $x_i$, we let $x_i^+$ and $x_i^-$ denote the mined positive and negative example, respectively. The parameter shared dual-tower encoder with different dropout is adopted to produce the sentence embedding, just as SimCSE does. However, $x_i$ and $x_i^+$ are used as positive example to be feed into the dual-tower encoder as input.

Obtaining two sentence embedding $ \mathbf{h}_{x_i}$ and $ \mathbf{h}_{x_i^+}$ produced by encoder from two positive examples , we compute their normalized representation $ \mathbf{z}_{x_i}$  and $\mathbf{z}_{x_i^+}$ leveraging batch normalization \cite{ioffe2015batch}. Then we setup an alternative prediction problem with the following loss function:

\begin{equation}
	\mathcal{L}(h_{x_i},h_{x_i^+}) =  {L}(z_{x_i},h_{x_i^+}) +   {L}(z_{x_i^+},h_{x_i})
\end{equation}

This loss function contains two terms that setup the alternative prediction problem of predicting the normalized representation $ \mathbf{z}_{x_i}$ from the positive example's sentence embedding $\mathbf{h}_{x_i^+}$, and $\mathbf{z}_{x_i^+}$ from $ \mathbf{h}_{x_i}$. Each term represents the Debias-infoNCE loss as follows:

\begin{equation}
	{L}(z_{x_i},h_{x_i^+})= -log\frac{e^{{cos(\mathbf{z}_{x_i},\mathbf{h}_{x_i^+})}/\tau}}{\sum_{k=1}^{m}{e}^{cos(\mathbf{z}_{x_i},\mathbf{z}_{x_i^-})/\tau}+\sum_{j=1}^{N-m}e^{cos(\mathbf{z}_{x_i},\mathbf{z}_j)/\tau}}
\end{equation}
and
\begin{equation}
	{L}(z_{x_i^+},h_{x_i})= -log\frac{e^{{cos(\mathbf{z}_{x_i^+},\mathbf{h}_{x_i})}/\tau}}{ \sum_{k=1}^{m} {e}^{cos(\mathbf{z}_{x_i^+},\mathbf{z}_{x_i^-})/\tau}+\sum_{j=1}^{N-m}e^{cos(\mathbf{z}_{x_i^+},\mathbf{z}_j)/\tau}}
\end{equation}
where $ \mathbf{\tau}$ is temperature parameter and we set $ \mathbf{\tau=0.05}$ as default configurations of SimCSE does. 
$ {N}$ is the size of mini-batch and $ \mathbf{m}$ is the number of the mined hard negative examples.

Compared with infoNCE used by SimCSE, Debias-infoNCE loss replaces the same input sentence with the mined positive example and introduces hard negative examples into random negatives. This reduces the training data distribution gap between the contrastive and supervised learning and helps produce high-quality sentence representation.

\section{Experiments}

\subsection{Setups}
\paragraph{\textbf{Evaluation Tasks}}
Consistent with previous works \cite{gao2021simcse}, this work uses the SentEval toolkit \cite{conneau2018senteval} and semantic textual similarity (STS) datasets for evaluating unsupervised sentence embedding models. The STS dataset includes the following tasks: STS tasks, 2012 - 2016 \cite{agirre-etal-2012-semeval,agirre-etal-2013-sem,agirre2014semeval,agirre2015semeval,agirre2016semeval}, STS benchmark (STSb) \cite{cer2017semeval} and SICK Relatedness (SICK-R) \cite{marelli2014sick}.  STS dataset is one of the most widely used benchmark for the evaluation of unsupervised sentence embedding and aims to measure the semantic similarity of sentence pairs and tests the semantic representation ability of models. Each sample in the STS dataset contains a pair of sentences and a corresponding similarity score, which ranges from 0 to 5 and is used to measure the similarity between sentence pair. 

\paragraph{\textbf{Baselines}} 
To verify the effectiveness of DebCSE, we compare our model with strong unsupervised baselines, including SimCSE \cite{gao2021simcse}, ArcCSE \cite{zhang2022contrastive}, DiffCSE \cite{chuang2022diffcse}, SNCSE \cite{wang2022sncse}, ESimCSE \cite{wu-etal-2022-esimcse},  BERT-flow \cite{li2020sentence}, ConSERT \cite{yan-etal-2021-consert}, IS-BERT \cite{zhang2020unsupervised}, and Bert-whitening \cite{su2021whitening}. Some baseline methods are evaluated on both the base and large parameter scales, while others are only evaluated on the base parameter scale..

\paragraph{\textbf{Implementation Details}} 
We use Wikipedia-1M data as the training dataset,  which is randomly sampled 1,000,000 sentences from Wikipedia. During training, we train our models for 1 epoch with temperature $\tau$ = 0.05 using an Adam optimizer. The batch size is 64 and the learning rate is 2.5e-5 and 5e-6  respectively, for base and large parameter scale models.  For the four backbone models, the dropout rate is set to 0.1.    Those models start from the pre-trained checkpoints of BERT-uncased \cite{devlin2018bert}  and RoBERTa-cased \cite{liu2019roberta} on Wiki-1M.

When constructing negative candicates pool, we use a pre-trained unsupervised SimCSE model with similarity score range between 0.25 and 0.75 as a semantic sentence filter to find out  those sentences which have not too low and not too high semantic similarity with the anchor sentence.

Following the evaluation process of SimCSE \cite{gao2021simcse} for a fair comparison, we evaluate the model every 125 training steps on the development set of STS-B and SICK-R to keep the best checkpoint for the final evaluation on test sets. We implement our framework with Huggingface’s transformers \cite{wolf-etal-2020-transformers}.

\subsection{Main Results}
Table {2} shows different approaches’ performances on seven semantic textual similarity (STS) test sets. It can be seen that DebCSE outperforms other SOTA methods substantially and achieves the best performance  in different parameter scale settings. Specifically,  DebCSE achieves on average 4.1 points absolute improvements in terms of Spearman’s correlation on BERTbase compared with SimCSE. Meanwhile, DebCSE obtains the best performance with the highest average Spearman’s correlation coefficient on all backbone encoders, which are respectively 80.33\% for BERT-base, 81.11\% for BERT-large, 80.60\% for RoBERTa-base and 82.01\% for RoBERTa-large. 

These experimental results demonstrate that DebCSE has better generalization ability over SOTA unsupervised sentence embedding methods and prove that reducing the "training data distribution bias" is indeed the key problem for sentence embedding learning.

\begin{figure*}[htbp]
	\begin{minipage}[t]{0.5\linewidth}
		\centering
		\includegraphics[width=\textwidth]{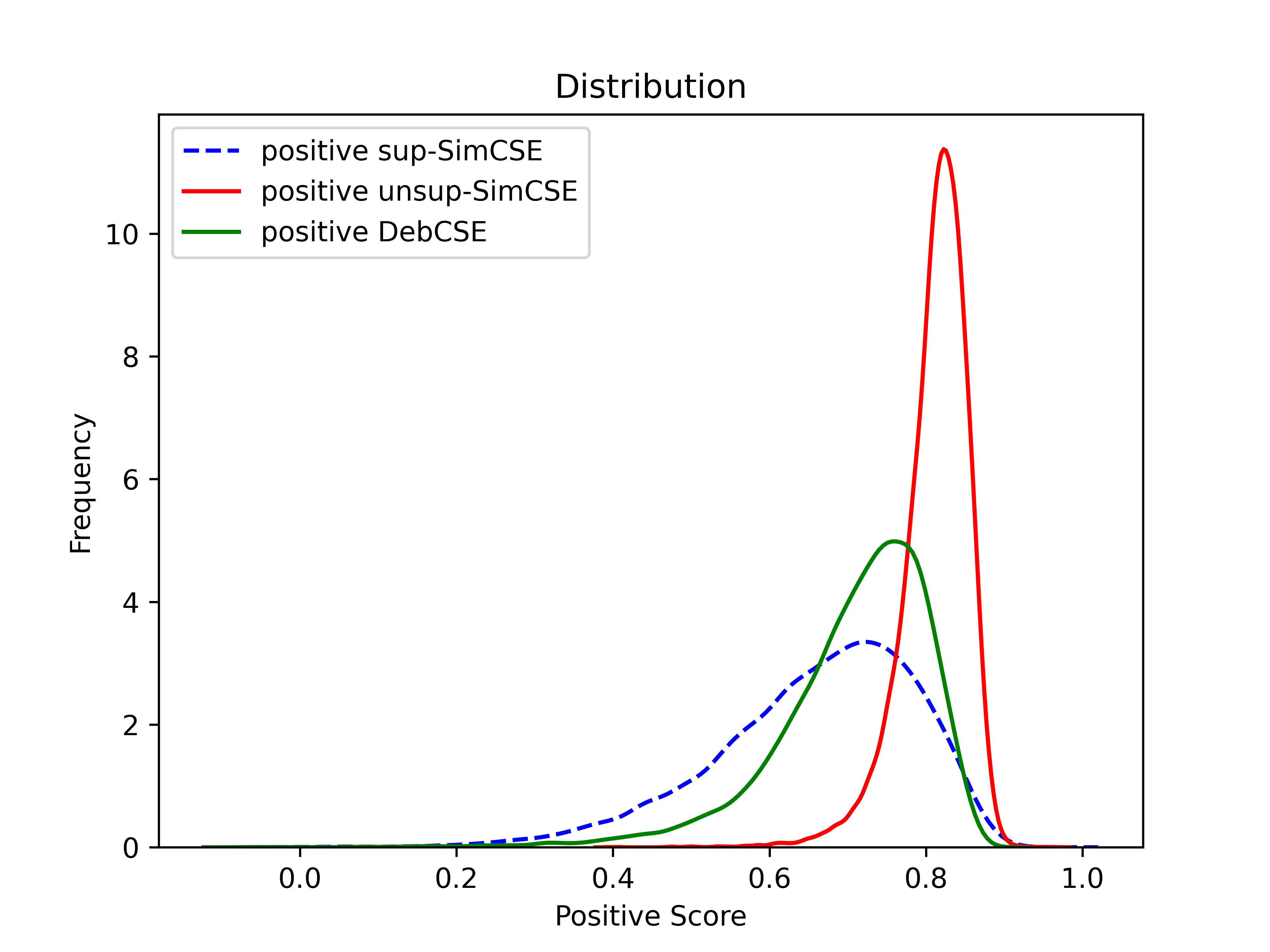}
		\centerline{(a) Similarity distribution of positive pairs }
	\end{minipage}%
	\begin{minipage}[t]{0.5\linewidth}
		\centering
		\includegraphics[width=\textwidth]{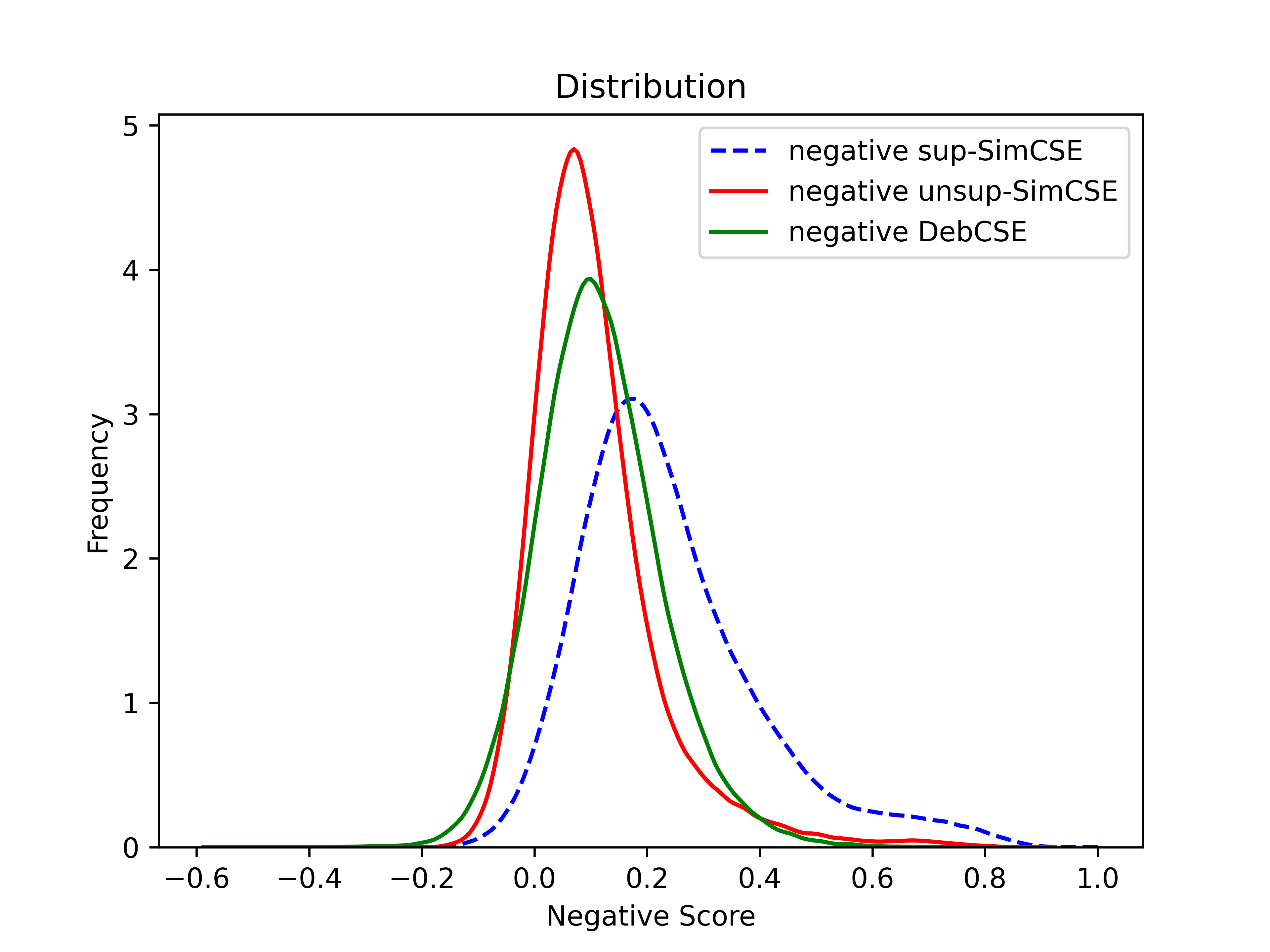}
		\centerline{(b) Similarity distribution of negative pairs}
	\end{minipage}
	\caption{Effectiveness of Debias(Semantic)}
\end{figure*}

\section{Analysis}
In this section, we firstly perform an extensive ablation studies to DebCSE with BERTbase on the development set of STS-B. Then, we further analyze the effect of DebCSE on reducing training data distribution bias. We also explore the distribution of sentence embedding based on Uniformity and Alignment, and the effect of hyper-parameters.

\subsection{Ablation Studies}

\begin{table}[htbp]
	\centering
	\caption{Ablation study of different components}
	\label{table3}%
	\begin{tabular}{p{10em}c}
		\toprule
		\textbf{Method} & {\textbf{STS Avg.}} \\
		\midrule
		DebCSE & 80.33 \\
		\quad w/o Alternative Norm& 79.43 \\
		\quad w/o  Negative-Debias& 78.96 \\
		\quad w/o  Postive-Debias & 78.82 \\
		\bottomrule
	\end{tabular}

\end{table}%

To verify each component's contribution to the DebCSE's final result, we conduct ablation study by removing each of the components on seven STS tasks and report the average value of the Spearman’s correlation metric. As shown in Table 3, removing any component would lead to the sharp performance degradation of DebCSE. It indicates that all the three key components are important in our proposed framework. 

\subsection{Analysis of Training Data Distribution Bias}

\begin{table}[htbp]
	\centering
	\caption{Effectiveness of Debias(Surface)}
	\label{table4}%
	\begin{tabular}{cccc}
		\toprule
		\textbf{} & {\textbf{Supervised}} & {\textbf{Contrastive}} & {\textbf{DebCSE}} \\
		\midrule
		Positive & 0.38  & 1.0 & \textbf{0.56}\\
		Negative  & 0.28   & 0.09 & \textbf{0.21}\\
		\bottomrule
	\end{tabular}

\end{table}%

In this section we try to explore the effect on training data distribution bias after introducing our proposed DebCSE.

We use the mined positive and negative pairs as described in Section 4 as the training data of DebCSE and compare both the training data surface similarity distribution and semantic similarity distribution of DebCSE with that of commonly used contrastive and supervised learning.

We compute the lexical overlap between the positive pairs or negative pairs as surface similarity of training data and show the results in Table {4}. It can be seen that the surface similarity of both positive and negative pairs of DebCSE are between the corresponding values of contrastive and supervised learning. This demonstrates that the training data of DebCSE is much harder and informative than that of contrastive sentence learning method while DebCSE indeed alleviates the surface training data distribution bias problem.

We also use a pre-trained SimCSE as sentence encoder and cosine function as similarity metric to measure semantic similarity of positive and negative pairs. Figure 4 shows the comparison results of three approaches. We can see that distribution of training data of DebCSE is near the corresponding distribution of supervised learning, compared with the training data distribution of contrastive learning, which demonstrates DebCSE reduces the semantic training data distribution gap between contrastive and supervised learning bias.

\subsection{Uniformity and Alignment}
The quality of learned representations can also be measured by Uniformity and Alignment. Alignment is computed based on the expected distance between representations of positive pairs. Lower alignment indicates close representations between positive pairs.  Besides, Uniformity measures how well representations are uniformly scattered across the representation space.

\begin{table}[htbp]
	\centering
	\caption{Uniformity and Alignment}
	\label{table5}%
	\begin{tabular}{cccc}
		\toprule
		\textbf{} &  {\textbf{unsup-SimCSE}} & {\textbf{sup-SimCSE}}& {\textbf{DebCSE}}  \\
		\midrule
		
		Uniformity   &-2.26 & -1.13 & \textbf{-1.61}\\
		Alignment     & 0.38 & 0.16 & \textbf{0.24}\\
		\bottomrule
	\end{tabular}
\end{table}%

We follow prior studies \cite{wang2020understanding,gao2021simcse} to measure Uniformity on the whole STS-B test set, and Alignment on sentence pairs in STS-B test set with correlation scores higher than 4. As shown in Table 5, DebCSE shows better Alignment and higher Uniformity than the unsupervised SimCSE model based on Bert-Base. We conjecture that DebCSE obtains higher uniformity and better Alignment because of the sampled hard negative and positive pairs, which makes DebCSE behave more like supervised contrastive models.
Compared to the supervised SimCSE model, DebCSE achieves lower Uniformity but larger Alignment and we think that DebCSE obtains lower uniformity by still adopting random in-batch negatives.

\subsection{Hyper-parameters Analysis} 


\begin{table}[htbp]
	\centering
	\caption{Hyper-parameters Analysis Result}
	\label{table6}%
	\resizebox{\linewidth}{!}{
		\begin{tabular}{c|cccccc}
			\hline
			\multicolumn{1}{p{3.5em}}{\diagbox{${\lambda_{n}}$}{${\lambda_{p}}$}} & 1     & 0.8   & 0.6   & 0.4   & 0.2   & 0 \\
			\hline
			1     & \cellcolor[rgb]{ .843,  .929,  .875}79.46  & \cellcolor[rgb]{ .6,  .831,  .663}79.93  & \cellcolor[rgb]{ .49,  .788,  .569}80.14  & \cellcolor[rgb]{ .749,  .894,  .796}79.64  & \cellcolor[rgb]{ .506,  .792,  .584}80.11  & \cellcolor[rgb]{ .961,  .976,  .976}79.24  \\
			0.8   & \cellcolor[rgb]{ .639,  .847,  .702}79.85  & \cellcolor[rgb]{ .388,  .745,  .482}80.33  & \cellcolor[rgb]{ .584,  .824,  .651}79.96  & \cellcolor[rgb]{ .776,  .902,  .816}79.59  & \cellcolor[rgb]{ .8,  .914,  .839}79.55  & \cellcolor[rgb]{ .89,  .949,  .918}79.37  \\
			0.6   & \cellcolor[rgb]{ .498,  .792,  .58}80.12  & \cellcolor[rgb]{ .431,  .765,  .522}80.25  & \cellcolor[rgb]{ .443,  .769,  .529}80.23  & \cellcolor[rgb]{ .494,  .788,  .573}80.13  & \cellcolor[rgb]{ .475,  .78,  .557}80.17  & \cellcolor[rgb]{ .588,  .827,  .655}79.95  \\
			0.4   & \cellcolor[rgb]{ .608,  .835,  .675}79.91  & \cellcolor[rgb]{ .608,  .835,  .675}79.91  & \cellcolor[rgb]{ .435,  .765,  .525}80.24  & \cellcolor[rgb]{ .525,  .804,  .6}80.07  & \cellcolor[rgb]{ .439,  .769,  .525}80.24  & \cellcolor[rgb]{ .918,  .961,  .941}79.32  \\
			0.2   & \cellcolor[rgb]{ .914,  .957,  .933}79.33  & \cellcolor[rgb]{ .718,  .878,  .769}79.70  & \cellcolor[rgb]{ .498,  .792,  .58}80.12  & \cellcolor[rgb]{ .741,  .89,  .784}79.66  & \cellcolor[rgb]{ .533,  .804,  .608}80.06  & \cellcolor[rgb]{ .98,  .988,  .996}79.20  \\
			0     & \cellcolor[rgb]{ .945,  .973,  .961}79.27  & \cellcolor[rgb]{ .827,  .925,  .863}79.49  & \cellcolor[rgb]{ .761,  .898,  .804}79.62  & \cellcolor[rgb]{ .902,  .953,  .925}79.35  & \cellcolor[rgb]{ .812,  .918,  .847}79.53  & \cellcolor[rgb]{ .988,  .988,  1}79.18  \\
			\bottomrule
		\end{tabular}%
	}

\end{table}%
In this section, we conduct extensive experiments to explore the effects of weighting hyper-parameter of semantic similarity $ \lambda_p$ and  $ \lambda_n$ on model's performance. 

Here $ \lambda_p$ denotes weighting parameter of semantic similarity when we sample positive examples while $ \lambda_n$  means weighting parameter of semantic similarity when we sample negative examples. Setting the parameter a larger value means the semantic similarity is more important than surface similarity during the sampling procedure. We adopt various values for $ \lambda_p$ and $ \lambda_n$ and Table 6 shows the experimental results.

From the experimental results, we mainly have the following observations:

(1) When $ \lambda_p=0.8$ and $ \lambda_n=0.8$, we have the best performance model. This demonstrates the semantic similarity is more important than surface similarity during the training data sampling procedure, no matter positive or negative example sampling.

(2) When we only use semantic similarity and ignore the effect of surface similarity($ \lambda_p=1$ or $ \lambda_n=1$), model performance is sub-optimal. This indicates surface similarity is beneficial to model performance.

All in all, we can draw the conclusion that surface similarity and semantic similarity both contribute to the final performance gain while semantic similarity is more important during the training data sampling procedure.

\section{Related Work}

Both supervised and unsupervised learning have been widely studied in the field of sentence embedding learning. Early supervised sentence embedding learning studies such as \cite{reimers2019sentence} mainly focus on supervised learning methods based on Natural Language Inference. Unsupervised approaches such as \cite{mikolov2013distributed} try to use pooling word2vec embeddings to derive representations without labeled datasets. A plethora of studies, such as \cite{reimers2019sentence,li2020sentence,yan-etal-2021-consert} have established that native sentence embedding generated directly from pre-trained language models lack the desired level of quality. And it's embeddings is often highly similar and concentrated, which are unable to accurately reflect the semantic similarity and relatedness between sentences. According to the \cite{li2020sentence}, word frequency bias is an important cause of the poor performance of sentence embedding and lead to the collapse and concentration of sentence vectors due to the uneven distribution of word frequencies in the representation space. To address this issue, it proposes a method that aims to improve the performance of the BERT model by reducing word frequency bias through the use of a flow model.  While \cite{su2021whitening} introduced whitening method to resolved the problem.

In recent years, contrastive learning methods such as SimCSE\cite{gao2021simcse} and ConSERT \cite{yan-etal-2021-consert} have achieved great success in unsupervised sentence representation learning. Contrastive learning is based on the principle of maximizing the mutual information between different views and has demonstrated to be highly effective in learning robust sentence representations in an unsupervised setting. For example, SimCSE use dropout as data augment function to make positive examples to obtain high-quality sentence representations. ConSERT \cite{yan-etal-2021-consert} explores various effective data augmentation strategies, and ESimCSE \cite{wu2021esimcse} attempts to attenuate the length bias by randomly repeating words. \cite{zhou2022debiased} try to reduce the sampling biased cause by improper negatives pairs. Besides, ArcCSE \cite{zhang2022contrastive} design triplet loss to enhance the pairwise discriminative ability of the model. DiffCSE \cite{chuang2022diffcse} learns sentence embeddings that are sensitive to the difference between the original sentences and an edited sentences, which improves the quality of representation. While SNCSE \cite{wang2022sncse} proposes contrastive learning for unsupervised sentence embedding with soft negative samples. 

Different from all the current SOTA methods, we regard the contrastive sentence learning as a debiasing procedure, which tries to fill the training data distribution gap between the contrastive learning and supervised learning. \textbf{It's not hard to see that many of current SOTA approaches are just a special case of our proposed DebCSE framework, which try to reduce either the positive example bias or negative example bias introduced by the simple training data construction rules in contrastive sentence learning}.

\section{Conclusion and Future Work}
In this paper, we proposed the DebCSE, a debiased contrastive learning framework for unsupervised sentence representation learning. We think the simple rules used to produce training data in contrastive learning bring various biases and propose to sample positive and negative examples by inverse propensity weighting, which reduces the training data distribution gap between the contrastive learning and supervised learning. We demonstrate the effectiveness of our framework on seven STS tasks and experimental results have shown that our approach outperforms several competitive baselines. 

Additionally, we believe that using large language models such as ChatGPT to create positive and negative examples for contrastive learning tasks will greatly diminish the various biases present in the current contrastive learning training data.  In the future, we plan to adopt this strategy to enhance the representation quality of producing sentence embeddings by contrastive learning.

\balance
\bibliographystyle{ACM-Reference-Format}
\bibliography{sample-base}


\begin{thebibliography}{39}


\ifx \showCODEN    \undefined \def \showCODEN     #1{\unskip}     \fi
\ifx \showDOI      \undefined \def \showDOI       #1{#1}\fi
\ifx \showISBNx    \undefined \def \showISBNx     #1{\unskip}     \fi
\ifx \showISBNxiii \undefined \def \showISBNxiii  #1{\unskip}     \fi
\ifx \showISSN     \undefined \def \showISSN      #1{\unskip}     \fi
\ifx \showLCCN     \undefined \def \showLCCN      #1{\unskip}     \fi
\ifx \shownote     \undefined \def \shownote      #1{#1}          \fi
\ifx \showarticletitle \undefined \def \showarticletitle #1{#1}   \fi
\ifx \showURL      \undefined \def \showURL       {\relax}        \fi
\providecommand\bibfield[2]{#2}
\providecommand\bibinfo[2]{#2}
\providecommand\natexlab[1]{#1}
\providecommand\showeprint[2][]{arXiv:#2}

\bibitem[Agirre et~al\mbox{.}(2015)]%
        {agirre2015semeval}
\bibfield{author}{\bibinfo{person}{Eneko Agirre}, \bibinfo{person}{Carmen
  Banea}, \bibinfo{person}{Claire Cardie}, \bibinfo{person}{Daniel Cer},
  \bibinfo{person}{Mona Diab}, \bibinfo{person}{Aitor Gonzalez-Agirre},
  \bibinfo{person}{Weiwei Guo}, \bibinfo{person}{Inigo Lopez-Gazpio},
  \bibinfo{person}{Montse Maritxalar}, \bibinfo{person}{Rada Mihalcea},
  {et~al\mbox{.}}} \bibinfo{year}{2015}\natexlab{}.
\newblock \showarticletitle{Semeval-2015 task 2: Semantic textual similarity,
  english, spanish and pilot on interpretability}. In
  \bibinfo{booktitle}{\emph{Proceedings of the 9th international workshop on
  semantic evaluation (SemEval 2015)}}. \bibinfo{pages}{252--263}.
\newblock


\bibitem[Agirre et~al\mbox{.}(2014)]%
        {agirre2014semeval}
\bibfield{author}{\bibinfo{person}{Eneko Agirre}, \bibinfo{person}{Carmen
  Banea}, \bibinfo{person}{Claire Cardie}, \bibinfo{person}{Daniel~M Cer},
  \bibinfo{person}{Mona~T Diab}, \bibinfo{person}{Aitor Gonzalez-Agirre},
  \bibinfo{person}{Weiwei Guo}, \bibinfo{person}{Rada Mihalcea},
  \bibinfo{person}{German Rigau}, {and} \bibinfo{person}{Janyce Wiebe}.}
  \bibinfo{year}{2014}\natexlab{}.
\newblock \showarticletitle{SemEval-2014 Task 10: Multilingual Semantic Textual
  Similarity.}. In \bibinfo{booktitle}{\emph{SemEval@ COLING}}.
  \bibinfo{pages}{81--91}.
\newblock


\bibitem[Agirre et~al\mbox{.}(2016)]%
        {agirre2016semeval}
\bibfield{author}{\bibinfo{person}{Eneko Agirre}, \bibinfo{person}{Carmen
  Banea}, \bibinfo{person}{Daniel Cer}, \bibinfo{person}{Mona Diab},
  \bibinfo{person}{Aitor Gonzalez~Agirre}, \bibinfo{person}{Rada Mihalcea},
  \bibinfo{person}{German Rigau~Claramunt}, {and} \bibinfo{person}{Janyce
  Wiebe}.} \bibinfo{year}{2016}\natexlab{}.
\newblock \showarticletitle{Semeval-2016 task 1: Semantic textual similarity,
  monolingual and cross-lingual evaluation}. In
  \bibinfo{booktitle}{\emph{SemEval-2016. 10th International Workshop on
  Semantic Evaluation; 2016 Jun 16-17; San Diego, CA. Stroudsburg (PA): ACL;
  2016. p. 497-511.}} ACL (Association for Computational Linguistics).
\newblock


\bibitem[Agirre et~al\mbox{.}(2012)]%
        {agirre-etal-2012-semeval}
\bibfield{author}{\bibinfo{person}{Eneko Agirre}, \bibinfo{person}{Daniel Cer},
  \bibinfo{person}{Mona Diab}, {and} \bibinfo{person}{Aitor Gonzalez-Agirre}.}
  \bibinfo{year}{2012}\natexlab{}.
\newblock \showarticletitle{{S}em{E}val-2012 Task 6: A Pilot on Semantic
  Textual Similarity}. In \bibinfo{booktitle}{\emph{*{SEM} 2012: The First
  Joint Conference on Lexical and Computational Semantics {--} Volume 1:
  Proceedings of the main conference and the shared task, and Volume 2:
  Proceedings of the Sixth International Workshop on Semantic Evaluation
  ({S}em{E}val 2012)}}. \bibinfo{publisher}{Association for Computational
  Linguistics}, \bibinfo{address}{Montr{\'e}al, Canada},
  \bibinfo{pages}{385--393}.
\newblock
\urldef\tempurl%
\url{https://aclanthology.org/S12-1051}
\showURL{%
\tempurl}


\bibitem[Agirre et~al\mbox{.}(2013)]%
        {agirre-etal-2013-sem}
\bibfield{author}{\bibinfo{person}{Eneko Agirre}, \bibinfo{person}{Daniel Cer},
  \bibinfo{person}{Mona Diab}, \bibinfo{person}{Aitor Gonzalez-Agirre}, {and}
  \bibinfo{person}{Weiwei Guo}.} \bibinfo{year}{2013}\natexlab{}.
\newblock \showarticletitle{*{SEM} 2013 shared task: Semantic Textual
  Similarity}. In \bibinfo{booktitle}{\emph{Second Joint Conference on Lexical
  and Computational Semantics (*{SEM}), Volume 1: Proceedings of the Main
  Conference and the Shared Task: Semantic Textual Similarity}}.
  \bibinfo{publisher}{Association for Computational Linguistics},
  \bibinfo{address}{Atlanta, Georgia, USA}, \bibinfo{pages}{32--43}.
\newblock
\urldef\tempurl%
\url{https://aclanthology.org/S13-1004}
\showURL{%
\tempurl}


\bibitem[Ai et~al\mbox{.}(2018)]%
        {ai2018unbiased}
\bibfield{author}{\bibinfo{person}{Qingyao Ai}, \bibinfo{person}{Keping Bi},
  \bibinfo{person}{Cheng Luo}, \bibinfo{person}{Jiafeng Guo}, {and}
  \bibinfo{person}{W~Bruce Croft}.} \bibinfo{year}{2018}\natexlab{}.
\newblock \showarticletitle{Unbiased learning to rank with unbiased propensity
  estimation}. In \bibinfo{booktitle}{\emph{The 41st international ACM SIGIR
  conference on research \& development in information retrieval}}.
  \bibinfo{pages}{385--394}.
\newblock


\bibitem[Cer et~al\mbox{.}(2017)]%
        {cer2017semeval}
\bibfield{author}{\bibinfo{person}{Daniel Cer}, \bibinfo{person}{Mona Diab},
  \bibinfo{person}{Eneko Agirre}, \bibinfo{person}{Inigo Lopez-Gazpio}, {and}
  \bibinfo{person}{Lucia Specia}.} \bibinfo{year}{2017}\natexlab{}.
\newblock \showarticletitle{Semeval-2017 task 1: Semantic textual
  similarity-multilingual and cross-lingual focused evaluation}.
\newblock \bibinfo{journal}{\emph{arXiv preprint arXiv:1708.00055}}
  (\bibinfo{year}{2017}).
\newblock


\bibitem[Chen and He(2021)]%
        {chen2021exploring}
\bibfield{author}{\bibinfo{person}{Xinlei Chen} {and} \bibinfo{person}{Kaiming
  He}.} \bibinfo{year}{2021}\natexlab{}.
\newblock \showarticletitle{Exploring simple siamese representation learning}.
  In \bibinfo{booktitle}{\emph{Proceedings of the IEEE/CVF conference on
  computer vision and pattern recognition}}. \bibinfo{pages}{15750--15758}.
\newblock


\bibitem[Chuang et~al\mbox{.}(2020)]%
        {chuang2020debiased}
\bibfield{author}{\bibinfo{person}{Ching-Yao Chuang}, \bibinfo{person}{Joshua
  Robinson}, \bibinfo{person}{Yen-Chen Lin}, \bibinfo{person}{Antonio
  Torralba}, {and} \bibinfo{person}{Stefanie Jegelka}.}
  \bibinfo{year}{2020}\natexlab{}.
\newblock \showarticletitle{Debiased contrastive learning}.
\newblock \bibinfo{journal}{\emph{Advances in neural information processing
  systems}}  \bibinfo{volume}{33} (\bibinfo{year}{2020}),
  \bibinfo{pages}{8765--8775}.
\newblock


\bibitem[Chuang et~al\mbox{.}(2022)]%
        {chuang2022diffcse}
\bibfield{author}{\bibinfo{person}{Yung-Sung Chuang}, \bibinfo{person}{Rumen
  Dangovski}, \bibinfo{person}{Hongyin Luo}, \bibinfo{person}{Yang Zhang},
  \bibinfo{person}{Shiyu Chang}, \bibinfo{person}{Marin Soljacic},
  \bibinfo{person}{Shang-Wen Li}, \bibinfo{person}{Wen-tau Yih},
  \bibinfo{person}{Yoon Kim}, {and} \bibinfo{person}{James Glass}.}
  \bibinfo{year}{2022}\natexlab{}.
\newblock \showarticletitle{{DiffCSE}: Difference-based Contrastive Learning
  for Sentence Embeddings}. In \bibinfo{booktitle}{\emph{Annual Conference of
  the North American Chapter of the Association for Computational Linguistics
  (NAACL)}}.
\newblock


\bibitem[Conneau and Kiela(2018)]%
        {conneau2018senteval}
\bibfield{author}{\bibinfo{person}{Alexis Conneau} {and} \bibinfo{person}{Douwe
  Kiela}.} \bibinfo{year}{2018}\natexlab{}.
\newblock \showarticletitle{Senteval: An evaluation toolkit for universal
  sentence representations}.
\newblock \bibinfo{journal}{\emph{arXiv preprint arXiv:1803.05449}}
  (\bibinfo{year}{2018}).
\newblock


\bibitem[Devlin et~al\mbox{.}(2018)]%
        {devlin2018bert}
\bibfield{author}{\bibinfo{person}{Jacob Devlin}, \bibinfo{person}{Ming-Wei
  Chang}, \bibinfo{person}{Kenton Lee}, {and} \bibinfo{person}{Kristina
  Toutanova}.} \bibinfo{year}{2018}\natexlab{}.
\newblock \showarticletitle{Bert: Pre-training of deep bidirectional
  transformers for language understanding}.
\newblock \bibinfo{journal}{\emph{arXiv preprint arXiv:1810.04805}}
  (\bibinfo{year}{2018}).
\newblock


\bibitem[Gao et~al\mbox{.}(2018)]%
        {gao2018representation}
\bibfield{author}{\bibinfo{person}{Jun Gao}, \bibinfo{person}{Di He},
  \bibinfo{person}{Xu Tan}, \bibinfo{person}{Tao Qin}, \bibinfo{person}{Liwei
  Wang}, {and} \bibinfo{person}{Tieyan Liu}.} \bibinfo{year}{2018}\natexlab{}.
\newblock \showarticletitle{Representation Degeneration Problem in Training
  Natural Language Generation Models}. In
  \bibinfo{booktitle}{\emph{International Conference on Learning
  Representations}}.
\newblock


\bibitem[Gao et~al\mbox{.}(2021)]%
        {gao2021simcse}
\bibfield{author}{\bibinfo{person}{Tianyu Gao}, \bibinfo{person}{Xingcheng
  Yao}, {and} \bibinfo{person}{Danqi Chen}.} \bibinfo{year}{2021}\natexlab{}.
\newblock \showarticletitle{Simcse: Simple contrastive learning of sentence
  embeddings}.
\newblock \bibinfo{journal}{\emph{arXiv preprint arXiv:2104.08821}}
  (\bibinfo{year}{2021}).
\newblock


\bibitem[Grill et~al\mbox{.}(2020)]%
        {grill2020bootstrap}
\bibfield{author}{\bibinfo{person}{Jean-Bastien Grill},
  \bibinfo{person}{Florian Strub}, \bibinfo{person}{Florent Altch{\'e}},
  \bibinfo{person}{Corentin Tallec}, \bibinfo{person}{Pierre Richemond},
  \bibinfo{person}{Elena Buchatskaya}, \bibinfo{person}{Carl Doersch},
  \bibinfo{person}{Bernardo Avila~Pires}, \bibinfo{person}{Zhaohan Guo},
  \bibinfo{person}{Mohammad Gheshlaghi~Azar}, {et~al\mbox{.}}}
  \bibinfo{year}{2020}\natexlab{}.
\newblock \showarticletitle{Bootstrap your own latent-a new approach to
  self-supervised learning}.
\newblock \bibinfo{journal}{\emph{Advances in neural information processing
  systems}}  \bibinfo{volume}{33} (\bibinfo{year}{2020}),
  \bibinfo{pages}{21271--21284}.
\newblock


\bibitem[Ioffe and Szegedy(2015)]%
        {ioffe2015batch}
\bibfield{author}{\bibinfo{person}{Sergey Ioffe} {and}
  \bibinfo{person}{Christian Szegedy}.} \bibinfo{year}{2015}\natexlab{}.
\newblock \showarticletitle{Batch normalization: Accelerating deep network
  training by reducing internal covariate shift}. In
  \bibinfo{booktitle}{\emph{International conference on machine learning}}.
  pmlr, \bibinfo{pages}{448--456}.
\newblock


\bibitem[Jiang et~al\mbox{.}(2022)]%
        {jiang2022promptbert}
\bibfield{author}{\bibinfo{person}{Ting Jiang}, \bibinfo{person}{Shaohan
  Huang}, \bibinfo{person}{Zihan Zhang}, \bibinfo{person}{Deqing Wang},
  \bibinfo{person}{Fuzhen Zhuang}, \bibinfo{person}{Furu Wei},
  \bibinfo{person}{Haizhen Huang}, \bibinfo{person}{Liangjie Zhang}, {and}
  \bibinfo{person}{Qi Zhang}.} \bibinfo{year}{2022}\natexlab{}.
\newblock \showarticletitle{Promptbert: Improving bert sentence embeddings with
  prompts}.
\newblock \bibinfo{journal}{\emph{arXiv preprint arXiv:2201.04337}}
  (\bibinfo{year}{2022}).
\newblock


\bibitem[Joachims et~al\mbox{.}(2017)]%
        {joachims2017unbiased}
\bibfield{author}{\bibinfo{person}{Thorsten Joachims}, \bibinfo{person}{Adith
  Swaminathan}, {and} \bibinfo{person}{Tobias Schnabel}.}
  \bibinfo{year}{2017}\natexlab{}.
\newblock \showarticletitle{Unbiased learning-to-rank with biased feedback}. In
  \bibinfo{booktitle}{\emph{Proceedings of the tenth ACM international
  conference on web search and data mining}}. \bibinfo{pages}{781--789}.
\newblock


\bibitem[Kenton and Toutanova(2019)]%
        {kenton2019bert}
\bibfield{author}{\bibinfo{person}{Jacob Devlin Ming-Wei~Chang Kenton} {and}
  \bibinfo{person}{Lee~Kristina Toutanova}.} \bibinfo{year}{2019}\natexlab{}.
\newblock \showarticletitle{BERT: Pre-training of Deep Bidirectional
  Transformers for Language Understanding}. In
  \bibinfo{booktitle}{\emph{Proceedings of NAACL-HLT}}.
  \bibinfo{pages}{4171--4186}.
\newblock


\bibitem[Li et~al\mbox{.}(2020)]%
        {li2020sentence}
\bibfield{author}{\bibinfo{person}{Bohan Li}, \bibinfo{person}{Hao Zhou},
  \bibinfo{person}{Junxian He}, \bibinfo{person}{Mingxuan Wang},
  \bibinfo{person}{Yiming Yang}, {and} \bibinfo{person}{Lei Li}.}
  \bibinfo{year}{2020}\natexlab{}.
\newblock \showarticletitle{On the Sentence Embeddings from Pre-trained
  Language Models}. In \bibinfo{booktitle}{\emph{Proceedings of the 2020
  Conference on Empirical Methods in Natural Language Processing (EMNLP)}}.
  \bibinfo{pages}{9119--9130}.
\newblock


\bibitem[Liu et~al\mbox{.}(2019)]%
        {liu2019roberta}
\bibfield{author}{\bibinfo{person}{Yinhan Liu}, \bibinfo{person}{Myle Ott},
  \bibinfo{person}{Naman Goyal}, \bibinfo{person}{Jingfei Du},
  \bibinfo{person}{Mandar Joshi}, \bibinfo{person}{Danqi Chen},
  \bibinfo{person}{Omer Levy}, \bibinfo{person}{Mike Lewis},
  \bibinfo{person}{Luke Zettlemoyer}, {and} \bibinfo{person}{Veselin
  Stoyanov}.} \bibinfo{year}{2019}\natexlab{}.
\newblock \showarticletitle{Roberta: A robustly optimized bert pretraining
  approach}.
\newblock \bibinfo{journal}{\emph{arXiv preprint arXiv:1907.11692}}
  (\bibinfo{year}{2019}).
\newblock


\bibitem[Marelli et~al\mbox{.}(2014)]%
        {marelli2014sick}
\bibfield{author}{\bibinfo{person}{Marco Marelli}, \bibinfo{person}{Stefano
  Menini}, \bibinfo{person}{Marco Baroni}, \bibinfo{person}{Luisa Bentivogli},
  \bibinfo{person}{Raffaella Bernardi}, {and} \bibinfo{person}{Roberto
  Zamparelli}.} \bibinfo{year}{2014}\natexlab{}.
\newblock \showarticletitle{A SICK cure for the evaluation of compositional
  distributional semantic models}. In \bibinfo{booktitle}{\emph{Proceedings of
  the Ninth International Conference on Language Resources and Evaluation
  (LREC'14)}}. \bibinfo{pages}{216--223}.
\newblock


\bibitem[Mikolov et~al\mbox{.}(2013)]%
        {mikolov2013distributed}
\bibfield{author}{\bibinfo{person}{Tomas Mikolov}, \bibinfo{person}{Ilya
  Sutskever}, \bibinfo{person}{Kai Chen}, \bibinfo{person}{Greg~S Corrado},
  {and} \bibinfo{person}{Jeff Dean}.} \bibinfo{year}{2013}\natexlab{}.
\newblock \showarticletitle{Distributed representations of words and phrases
  and their compositionality}.
\newblock \bibinfo{journal}{\emph{Advances in neural information processing
  systems}}  \bibinfo{volume}{26} (\bibinfo{year}{2013}).
\newblock


\bibitem[Pennington et~al\mbox{.}(2014)]%
        {pennington2014glove}
\bibfield{author}{\bibinfo{person}{Jeffrey Pennington},
  \bibinfo{person}{Richard Socher}, {and} \bibinfo{person}{Christopher~D
  Manning}.} \bibinfo{year}{2014}\natexlab{}.
\newblock \showarticletitle{Glove: Global vectors for word representation}. In
  \bibinfo{booktitle}{\emph{Proceedings of the 2014 conference on empirical
  methods in natural language processing (EMNLP)}}.
  \bibinfo{pages}{1532--1543}.
\newblock


\bibitem[Reimers and Gurevych(2019)]%
        {reimers2019sentence}
\bibfield{author}{\bibinfo{person}{Nils Reimers} {and} \bibinfo{person}{Iryna
  Gurevych}.} \bibinfo{year}{2019}\natexlab{}.
\newblock \showarticletitle{Sentence-BERT: Sentence Embeddings using Siamese
  BERT-Networks}. In \bibinfo{booktitle}{\emph{Proceedings of the 2019
  Conference on Empirical Methods in Natural Language Processing and the 9th
  International Joint Conference on Natural Language Processing
  (EMNLP-IJCNLP)}}. \bibinfo{pages}{3982--3992}.
\newblock


\bibitem[Ristad and Yianilos(1998)]%
        {ristad1998learning}
\bibfield{author}{\bibinfo{person}{Eric~Sven Ristad} {and}
  \bibinfo{person}{Peter~N Yianilos}.} \bibinfo{year}{1998}\natexlab{}.
\newblock \showarticletitle{Learning string-edit distance}.
\newblock \bibinfo{journal}{\emph{IEEE Transactions on Pattern Analysis and
  Machine Intelligence}} \bibinfo{volume}{20}, \bibinfo{number}{5}
  (\bibinfo{year}{1998}), \bibinfo{pages}{522--532}.
\newblock


\bibitem[Su et~al\mbox{.}(2021)]%
        {su2021whitening}
\bibfield{author}{\bibinfo{person}{Jianlin Su}, \bibinfo{person}{Jiarun Cao},
  \bibinfo{person}{Weijie Liu}, {and} \bibinfo{person}{Yangyiwen Ou}.}
  \bibinfo{year}{2021}\natexlab{}.
\newblock \showarticletitle{Whitening sentence representations for better
  semantics and faster retrieval}.
\newblock \bibinfo{journal}{\emph{arXiv preprint arXiv:2103.15316}}
  (\bibinfo{year}{2021}).
\newblock


\bibitem[Wang et~al\mbox{.}(2022b)]%
        {wang2022sncse}
\bibfield{author}{\bibinfo{person}{Hao Wang}, \bibinfo{person}{Yangguang Li},
  \bibinfo{person}{Zhen Huang}, \bibinfo{person}{Yong Dou},
  \bibinfo{person}{Lingpeng Kong}, {and} \bibinfo{person}{Jing Shao}.}
  \bibinfo{year}{2022}\natexlab{b}.
\newblock \showarticletitle{SNCSE: Contrastive Learning for Unsupervised
  Sentence Embedding with Soft Negative Samples}.
\newblock \bibinfo{journal}{\emph{arXiv preprint arXiv:2201.05979}}
  (\bibinfo{year}{2022}).
\newblock


\bibitem[Wang and Isola(2020)]%
        {wang2020understanding}
\bibfield{author}{\bibinfo{person}{Tongzhou Wang} {and}
  \bibinfo{person}{Phillip Isola}.} \bibinfo{year}{2020}\natexlab{}.
\newblock \showarticletitle{Understanding contrastive representation learning
  through alignment and uniformity on the hypersphere}. In
  \bibinfo{booktitle}{\emph{International Conference on Machine Learning}}.
  PMLR, \bibinfo{pages}{9929--9939}.
\newblock


\bibitem[Wang et~al\mbox{.}(2022a)]%
        {wang2022improving}
\bibfield{author}{\bibinfo{person}{Wei Wang}, \bibinfo{person}{Liangzhu Ge},
  \bibinfo{person}{Jingqiao Zhang}, {and} \bibinfo{person}{Cheng Yang}.}
  \bibinfo{year}{2022}\natexlab{a}.
\newblock \showarticletitle{Improving Contrastive Learning of Sentence
  Embeddings with Case-Augmented Positives and Retrieved Negatives}. In
  \bibinfo{booktitle}{\emph{Proceedings of the 45th International ACM SIGIR
  Conference on Research and Development in Information Retrieval}}.
  \bibinfo{pages}{2159--2165}.
\newblock


\bibitem[Wang et~al\mbox{.}(2016)]%
        {wang2016learning}
\bibfield{author}{\bibinfo{person}{Xuanhui Wang}, \bibinfo{person}{Michael
  Bendersky}, \bibinfo{person}{Donald Metzler}, {and} \bibinfo{person}{Marc
  Najork}.} \bibinfo{year}{2016}\natexlab{}.
\newblock \showarticletitle{Learning to rank with selection bias in personal
  search}. In \bibinfo{booktitle}{\emph{Proceedings of the 39th International
  ACM SIGIR conference on Research and Development in Information Retrieval}}.
  \bibinfo{pages}{115--124}.
\newblock


\bibitem[Wolf et~al\mbox{.}(2020)]%
        {wolf-etal-2020-transformers}
\bibfield{author}{\bibinfo{person}{Thomas Wolf}, \bibinfo{person}{Lysandre
  Debut}, \bibinfo{person}{Victor Sanh}, \bibinfo{person}{Julien Chaumond},
  \bibinfo{person}{Clement Delangue}, \bibinfo{person}{Anthony Moi},
  \bibinfo{person}{Pierric Cistac}, \bibinfo{person}{Tim Rault},
  \bibinfo{person}{Rémi Louf}, \bibinfo{person}{Morgan Funtowicz},
  \bibinfo{person}{Joe Davison}, \bibinfo{person}{Sam Shleifer},
  \bibinfo{person}{Patrick von Platen}, \bibinfo{person}{Clara Ma},
  \bibinfo{person}{Yacine Jernite}, \bibinfo{person}{Julien Plu},
  \bibinfo{person}{Canwen Xu}, \bibinfo{person}{Teven~Le Scao},
  \bibinfo{person}{Sylvain Gugger}, \bibinfo{person}{Mariama Drame},
  \bibinfo{person}{Quentin Lhoest}, {and} \bibinfo{person}{Alexander~M. Rush}.}
  \bibinfo{year}{2020}\natexlab{}.
\newblock \showarticletitle{Transformers: State-of-the-Art Natural Language
  Processing}. In \bibinfo{booktitle}{\emph{Proceedings of the 2020 Conference
  on Empirical Methods in Natural Language Processing: System Demonstrations}}.
  \bibinfo{publisher}{Association for Computational Linguistics},
  \bibinfo{address}{Online}, \bibinfo{pages}{38--45}.
\newblock
\urldef\tempurl%
\url{https://www.aclweb.org/anthology/2020.emnlp-demos.6}
\showURL{%
\tempurl}


\bibitem[Wu et~al\mbox{.}(2021)]%
        {wu2021esimcse}
\bibfield{author}{\bibinfo{person}{Xing Wu}, \bibinfo{person}{Chaochen Gao},
  \bibinfo{person}{Liangjun Zang}, \bibinfo{person}{Jizhong Han},
  \bibinfo{person}{Zhongyuan Wang}, {and} \bibinfo{person}{Songlin Hu}.}
  \bibinfo{year}{2021}\natexlab{}.
\newblock \showarticletitle{ESimCSE: Enhanced Sample Building Method for
  Contrastive Learning of Unsupervised Sentence Embedding}.
\newblock \bibinfo{journal}{\emph{arXiv preprint arXiv:2109.04380}}
  (\bibinfo{year}{2021}).
\newblock


\bibitem[Wu et~al\mbox{.}(2022)]%
        {wu-etal-2022-esimcse}
\bibfield{author}{\bibinfo{person}{Xing Wu}, \bibinfo{person}{Chaochen Gao},
  \bibinfo{person}{Liangjun Zang}, \bibinfo{person}{Jizhong Han},
  \bibinfo{person}{Zhongyuan Wang}, {and} \bibinfo{person}{Songlin Hu}.}
  \bibinfo{year}{2022}\natexlab{}.
\newblock \showarticletitle{{ES}im{CSE}: Enhanced Sample Building Method for
  Contrastive Learning of Unsupervised Sentence Embedding}. In
  \bibinfo{booktitle}{\emph{Proceedings of the 29th International Conference on
  Computational Linguistics}}. \bibinfo{publisher}{International Committee on
  Computational Linguistics}, \bibinfo{address}{Gyeongju, Republic of Korea},
  \bibinfo{pages}{3898--3907}.
\newblock
\urldef\tempurl%
\url{https://aclanthology.org/2022.coling-1.342}
\showURL{%
\tempurl}


\bibitem[Yan et~al\mbox{.}(2021)]%
        {yan-etal-2021-consert}
\bibfield{author}{\bibinfo{person}{Yuanmeng Yan}, \bibinfo{person}{Rumei Li},
  \bibinfo{person}{Sirui Wang}, \bibinfo{person}{Fuzheng Zhang},
  \bibinfo{person}{Wei Wu}, {and} \bibinfo{person}{Weiran Xu}.}
  \bibinfo{year}{2021}\natexlab{}.
\newblock \showarticletitle{{C}on{SERT}: A Contrastive Framework for
  Self-Supervised Sentence Representation Transfer}. In
  \bibinfo{booktitle}{\emph{Proceedings of the 59th Annual Meeting of the
  Association for Computational Linguistics and the 11th International Joint
  Conference on Natural Language Processing (Volume 1: Long Papers)}}.
  \bibinfo{publisher}{Association for Computational Linguistics},
  \bibinfo{address}{Online}, \bibinfo{pages}{5065--5075}.
\newblock
\urldef\tempurl%
\url{https://doi.org/10.18653/v1/2021.acl-long.393}
\showDOI{\tempurl}


\bibitem[Yang et~al\mbox{.}(2019)]%
        {yang2019xlnet}
\bibfield{author}{\bibinfo{person}{Zhilin Yang}, \bibinfo{person}{Zihang Dai},
  \bibinfo{person}{Yiming Yang}, \bibinfo{person}{Jaime Carbonell},
  \bibinfo{person}{Russ~R Salakhutdinov}, {and} \bibinfo{person}{Quoc~V Le}.}
  \bibinfo{year}{2019}\natexlab{}.
\newblock \showarticletitle{Xlnet: Generalized autoregressive pretraining for
  language understanding}.
\newblock \bibinfo{journal}{\emph{Advances in neural information processing
  systems}}  \bibinfo{volume}{32} (\bibinfo{year}{2019}).
\newblock


\bibitem[Zhang et~al\mbox{.}(2020)]%
        {zhang2020unsupervised}
\bibfield{author}{\bibinfo{person}{Yan Zhang}, \bibinfo{person}{Ruidan He},
  \bibinfo{person}{Zuozhu Liu}, \bibinfo{person}{Kwan~Hui Lim}, {and}
  \bibinfo{person}{Lidong Bing}.} \bibinfo{year}{2020}\natexlab{}.
\newblock \showarticletitle{An unsupervised sentence embedding method by mutual
  information maximization}.
\newblock \bibinfo{journal}{\emph{arXiv preprint arXiv:2009.12061}}
  (\bibinfo{year}{2020}).
\newblock


\bibitem[Zhang et~al\mbox{.}(2022)]%
        {zhang2022contrastive}
\bibfield{author}{\bibinfo{person}{Yuhao Zhang}, \bibinfo{person}{Hongji Zhu},
  \bibinfo{person}{Yongliang Wang}, \bibinfo{person}{Nan Xu},
  \bibinfo{person}{Xiaobo Li}, {and} \bibinfo{person}{Binqiang Zhao}.}
  \bibinfo{year}{2022}\natexlab{}.
\newblock \showarticletitle{A Contrastive Framework for Learning Sentence
  Representations from Pairwise and Triple-wise Perspective in Angular Space}.
  In \bibinfo{booktitle}{\emph{Proceedings of the 60th Annual Meeting of the
  Association for Computational Linguistics (Volume 1: Long Papers)}}.
  \bibinfo{pages}{4892--4903}.
\newblock


\bibitem[Zhou et~al\mbox{.}(2022)]%
        {zhou2022debiased}
\bibfield{author}{\bibinfo{person}{Kun Zhou}, \bibinfo{person}{Beichen Zhang},
  \bibinfo{person}{Wayne~Xin Zhao}, {and} \bibinfo{person}{Ji-Rong Wen}.}
  \bibinfo{year}{2022}\natexlab{}.
\newblock \showarticletitle{Debiased Contrastive Learning of Unsupervised
  Sentence Representations}.
\newblock \bibinfo{journal}{\emph{arXiv preprint arXiv:2205.00656}}
  (\bibinfo{year}{2022}).
\newblock


\end{thebibliography}

\appendix

\end{document}